\documentclass[runningheads]{llncs}
\usepackage{graphicx}
\begin{document}
\title{Building Safe and Reliable AI systems for Safety Critical Tasks with Vision-Language Processing}

\author{Shuang Ao\inst{1}\orcidID{0000-0003-2648-3082}} 
%
\authorrunning{F. Author et al.}
%
\institute{The Open University, Walton Hall, Kents Hill, Milton Keynes MK7 6AA
\email{shuang.ao@open.ac.uk}\\
}
\maketitle              
\begin{abstract}
Although AI systems have been applied in various fields and achieved impressive performance, their safety and reliability are still a big concern. This is especially important for safety-critical tasks. One shared characteristic of these critical tasks is their risk sensitivity, where small mistakes can cause big consequences and even endanger life. There are several factors that could be guidelines for the successful deployment of AI systems in sensitive tasks: (i) failure detection and out-of-distribution (OOD) detection; (ii) overfitting identification; (iii) uncertainty  quantification for predictions; (iv) robustness to data perturbations. These factors are also challenges of current AI systems, which are major blocks for building safe and reliable AI. Specifically, the current AI algorithms are unable to identify common causes for failure detection. Furthermore, additional techniques are required to quantify the quality of predictions. All these contribute to inaccurate uncertainty quantification, which lowers trust in predictions. Hence obtaining accurate model uncertainty quantification and its further improvement are challenging. To address these issues, many techniques have been proposed, such as regularization methods and learning strategies. As vision and language are the most typical data type and have many open source benchmark datasets, this thesis will focus on vision-language data processing for tasks like classification, image captioning, and vision question answering. In this thesis, we aim to build a safeguard by further developing current techniques to ensure the accurate model uncertainty for safety-critical tasks. 

\end{abstract}

\section{Introduction}
Despite the impressive performance of AI algorithm in various fields, their safety and reliability is still a concern. Recent studies have achieved successful performance in areas like image~\cite{he2016deep} and text classification~\cite{vaswani2017attention}, object detection~\cite{liu2016ssd}, segmentation~\cite{li2018h}, image captioning~\cite{zhou2020unified}, visual question answer~\cite{li2020visual} and graph scene generation~\cite{yang2018graph}, and some tasks obtain near-perfect results. However, AI has not been fully deployed in sensitive fields like autonomous driving, medical diagnosing, or assistance for socially vulnerable groups. The major limitation lies in the lack of safeguard in these safety-critical tasks. One shared characteristic of these tasks is their risk sensitivity: it raises serious concern as the mistake by the AI algorithm can be expensive and even endanger human life. Guidelines from academic papers~\cite{ashmore2021assuring} and industry whitepapers~\cite{aptiv2019safety} for deploying AI systems for safety-critical tasks include: identifying common causes of failure detection and out-of-distribution (OOD) detection, identifying overfitting in training data, quantifying uncertainty in prediction, and making the model robust to data perturbations. However, these recommendations are not fully satisfied in specific tasks, leading to the limitation of deployment of AI systems in these fields. 
 
One serious limitation of current AI systems is that they tend to give the wrong prediction confidently~\cite{szegedy2016rethinking,muller2019does}. Humans feel uncertain when their decision is potentially wrong or ambiguous in the decision-making process, and AI systems are supposed to have similar behaviors. In the recent decade, the quality of network architectures has significantly improved by utilizing deeper and wider networks such as VGG~\cite{simonyan2014very} and ResNet~\cite{he2016deep}. These state-of-the-art networks significantly improve feature learning for text and image but also raise the question of models with poor uncertainty quantification and being over-confident. It refers to when the overall confidence score is higher than the overall accuracy for testing data. Specifically, the model is supposed to show low confidence when the prediction is ambiguous or likely to be wrong and vice versa. The over-confident issue leads to the concern of inaccurate uncertainty quantification and trustworthiness of predictions.

The confidence and accuracy level of the system should match so that human experts can tell when the system tends to make mistakes. Hence  addressing the over-confidence issue is essential to building a safe and reliable AI system. Even though a deep learning model can output the prediction for trained tasks, it cannot provide feedback about the quality of its prediction. For example, which class is poorly performed or if the overall output is reliable or not. In other words, such quality refers to how doubtful or uncertain the model for its prediction is, known as the model uncertainty. Ideally, when the model uncertainty is high, the model should suggest a second opinion and defer the task to human experts to re-examine it. With human intervention, the unexpected behaviour or wrong predictions from the AI system can be prevented. This process is crucial for safety-critical tasks as it can enhance failure detection, meaning a model to detect its own wrong predictions during the deployment or real-time applications without checking with the ground truth. Hence precise quantification and sufficient improvement of model uncertainty lie in the heart of building a safe and reliable AI.
 
In practice, many safety-critical tasks require multi-modality processing, especially with vision and language data. For example, autonomous driving systems process images, audio from user's input, and also signal data from sensors. Models for medical diagnosing deal with the image data such as MRI and text data of patients records. As real-world applications are more complicated than single modality data processing such as solely image or text classification, I will research reliable multi-modal data processing of image and text, and possibly other data sources.

\section{Research Question}

\label{chap:research_questions}

The main goal of my research is to build a safeguard for vision-language processing, by developing techniques and learning strategies to improve estimates of model uncertainty. RQ1.2 and RQ 2.1 will be discussed at the Doctoral Consortium.

\noindent \textit{\textbf{RQ1: Can model uncertainty quantification be improved without adding additional computational complexity to build safe AI systems?}}
\begin{itemize}
   \item \textbf{RQ1.1: Can we improve upon the uniform distribution in Label Smoothing by generalizing the soft label in a more reasonable way for different applications? }
    
    The traditional label smoothing clips the hard label into uniformly distributed soft labels, but not the case in practice. Hence we need to tackle this issue for a more accurate soft label generalization. 

    \item \textbf{RQ1.2: How to efficiently conduct automatic failure detection (FD) for identify model's own wrong prediction during inference time?}

    FD is a significant criteria for the trustworthiness of a model. 

    \item \textbf{RQ1.3: For curriculum learning, can we rank difficulty of samples more accurately to build the framework of curriculum learning?}
    
    We will use the model confidence as a proxy for ranking the difficulty of training samples. 
    
\end{itemize}

\noindent \textit{\textbf{RQ2: By integrating the techniques of safe AI systems developed in RQ1, can we improve the reliability and robustness of the model in the application of vision-language processing?}} 
\begin{itemize} 
    
    \item \textbf{RQ2.1: For vision-language processing, how to build an end-to-end pipeline to reduce the dependency on the prior procedure, as well as the computational cost? }
    
    Tasks like image captioning and VQA include several processing steps such as object detection and feature extraction. Hence it is necessary to reduce the dependency of the prior step, to reduce its influence to latter processing.
\end{itemize}

\begin{figure}[!h]
\centerline{\includegraphics[width=0.9\textwidth]{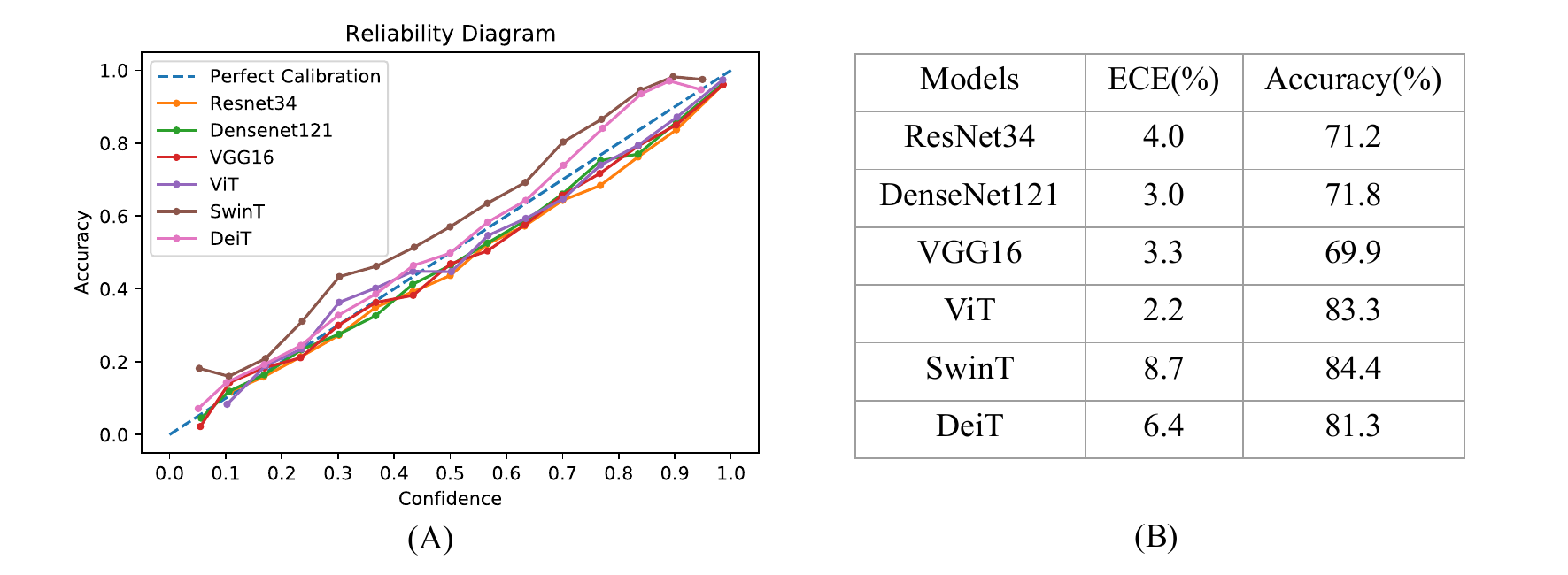}}

\caption{Left (A): Reliability diagram for ResNet34, DenseNet121, VGG16, ViT, SwinT, and DeiT models with ImageNet dataset. Right (B): Results of accuracy and ECE of various model architectures on ImageNet dataset. All results are shown in percentages. The lower the ECE the better.}
\label{fig:curve}
\end{figure}

\section{Preliminary Result}
We investigate 3 CNNs (ResNet34~\cite{he2016deep}, DenseNet121~\cite{huang2017densely} and VGG16~\cite{simonyan2014very}) and 3 transformers models (ViT~\cite{dosovitskiy2020image}, SwinT~\cite{liu2021swin} and Deit~\cite{touvron2021training}) for ImageNet~\cite{russakovsky2015imagenet} and CIFAR100~\cite{krizhevsky2009learning} dataset. The evaluation metrics are accuracy and Expected Calibration Error 
 (ECE)~\cite{naeini2015obtaining}, and ECE is the metric to measure calibration. In figure~\ref{fig:curve}, the table shows that models that perform better in accuracy are not always better in ECE, such as VGG16 and SwinT, which is illustrated further in the reliability diagram in ~\ref{fig:curve} (A). Hence it is essential to build a model with good calibration and high accuracy to improve the trustworthiness and reliability.

\begin{figure}[!h]
\centerline{\includegraphics[width=0.7\textwidth]{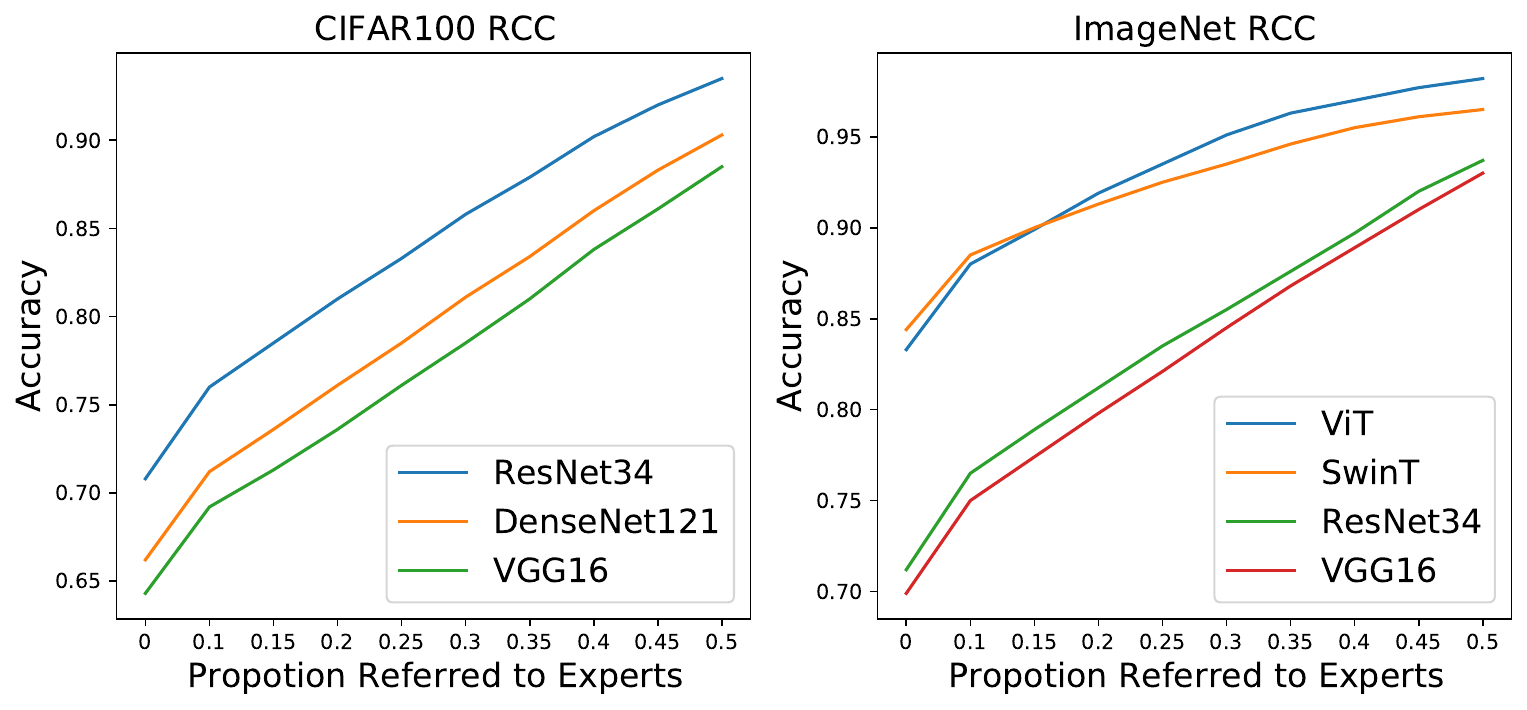}}

\caption{Risk Coverage Curve for CIFAR100 and ImageNet dataset of CNNs and transformers. The x-axis indicates the percentage of data removed from the entire test set, and the accuracy is calculated on the remaining test set. The higher accuracy means better performance of automatic failure detection.}
\label{fig:rcc}
\end{figure}

The Risk Coverage Curve (RCC) demonstrates the efficiency of automatic failure detection (AFD) as shown in Figure~\ref{fig:rcc}. We utilize the predictive uncertainty to distinguish correct from incorrect samples by following~\cite{corbiere2019addressing}. The data belonging to ``Referred to experts'' are wrongly predicted. Comparing ViT and SwinTrans models in the ImageNet RCC, SwinTrans has higher accuracy with the entire test set, but ViT outperforms it with more wrong predictions removed. It suggests that ViT is more reliable than SwinT as the model can detect more wrong predictions.

\section{Conclusion}
In this thesis, I will explore the topic of safe and reliable AI, focusing on the applications in vision-language processing, such as image captioning and visual question answering. The main goal of my research is to build a safeguard for safety-critical tasks with multi-modal data processing.

\pagebreak

\bibliographystyle{splncs04}
\bibliography{reference}

\end{document}